\documentclass[conference]{IEEEtran}
\IEEEoverridecommandlockouts

\usepackage{cite}
\usepackage{amsmath,amssymb}
\usepackage{graphicx}
\usepackage{xcolor}
\usepackage{booktabs}
\usepackage{multirow}
\usepackage{makecell}
\usepackage{array}
\usepackage{pifont}
\usepackage{url}
\usepackage[hidelinks]{hyperref}

\definecolor{plannedred}{RGB}{214,40,40}
\newif\iffinalcheck
\finalcheckfalse

\newcommand{\plannedpanel}[1]{\fbox{\parbox{0.9\linewidth}{\centering\textcolor{plannedred}{\small #1}}}}

\newcommand{\sys}{JARVIS}
\newcommand{\llmname}{Claude Opus~5.5}
\newcommand{\twin}{executable digital twin}

\newcommand{\teasercaption}{\textbf{From a need in plain words to a tested board and the next design, with no human
hands in between.} An embodied agent turns a need stated in plain language into a requirement with numbers, and the
requirement into a design for the circuit and its board. It rehearses the whole build in an executable digital twin of
the bench, which the agent built by scanning the bench and which runs the same programs as the real machines. It builds
the board on unmodified desktop machines, which print the traces in conductive silver ink, dispense solder paste, place
the parts and reflow the solder. It then tests the board
hardware-in-the-loop on an OPAL-RT real-time simulator. Finally, it analyzes what it measured, explains it, and turns
it into the next design, which it draws and rehearses. The Design tile shows the requirement that the agent set and
the agent's board layout; the Build tile shows
the whole board after each step, drawn from that layout; the Test tile is a schematic, with an oscilloscope as a second
instrument on the same signals.}

\newcommand{\cmark}{\ding{51}}
\newcommand{\xmark}{--}
\newcommand{\pmark}{$\circ$}

\newcommand{\figref}[1]{Fig.~\ref{#1}}
\newcommand{\tabref}[1]{Table~\ref{#1}}
\newcommand{\secref}[1]{Sec.~\ref{#1}}
\newcommand{\para}[1]{\noindent\textbf{#1}\ \ }

\graphicspath{{figures/}}

\makeatletter
\let\@oldmaketitle\@maketitle
\renewcommand{\@maketitle}{\@oldmaketitle
  \begin{center}
    \includegraphics[width=\textwidth]{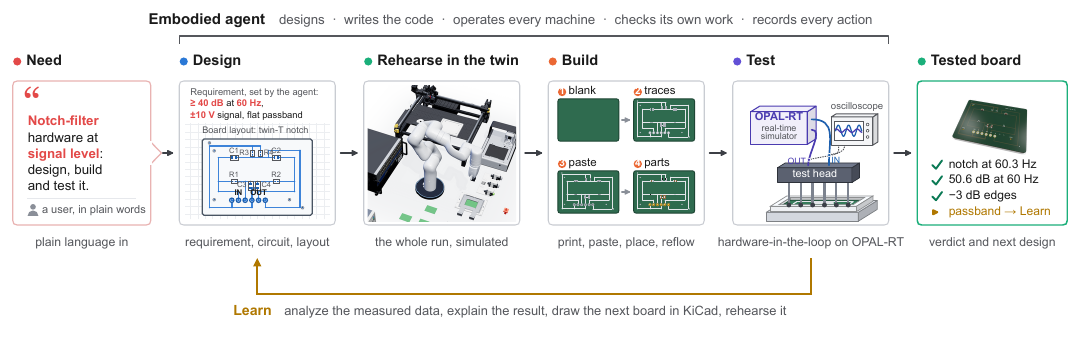}
  \end{center}
  \vspace{-0.6em}
  {\footnotesize Fig.~1.\ \ \teasercaption}
  \vspace{0.4em}}
\makeatother

\title{JARVIS: An Agent That Designs, Builds,\\and Tests Circuit Boards on Its Own}

\author{\IEEEauthorblockN{Shibo Zhou,~\IEEEmembership{Student~Member,~IEEE},
Tianqi Hong,~\IEEEmembership{Member,~IEEE},\\
Zheng Yan,~\IEEEmembership{Student~Member,~IEEE,},
and He Yang,~\IEEEmembership{Student~Member,~IEEE,}}
\IEEEauthorblockA{School of ECE, University of Georgia, Athens, GA, USA\\
Email: \{shibo.zhou; tianqi.hong; zheng.yan; He.Yang95\}@uga.edu}}

\begin{document}

\maketitle
\setcounter{figure}{1}

\begin{abstract}
Power electronics carries the energy transition, yet its hardware still improves at the speed of human hands: an
engineer must design, build, test and improve each new board. We present \sys{}, an embodied agent that does all
four by itself. \sys{} first knows its machines: for each one, it keeps a record of what the machine is, how to reach
it and what its limits are. It then scans its bench with a 3D scanner on its robot arm, obtains a panorama of the whole
bench and builds from it, on its own, an executable digital twin of the bench, in which the programs that drive the
real machines also run. In the twin, the agent rehearses the experiment before it uses any material: every motion of
the arm, the heating of the board, the duration of every step and the response of the circuit. One sentence typed on
a screen then starts the run. The agent builds the board on unmodified desktop machines. The arm carries the board and
changes the tool heads of a desktop PCB printer, which deposits conductive silver ink and solder paste and follows the
surface of the board as it prints. A pick-and-place machine sets the parts. The agent tests the board
hardware-in-the-loop on an OPAL-RT real-time simulator, with an oscilloscope as a second instrument, and
it checks every step of its work through its 3D scanner and the machines' own cameras and sensors. Finally, it analyzes the
measured data, explains where and why the board departs from its design, and proposes the next design, which it draws
in KiCad and rehearses in the twin. On a 60\,Hz notch filter of the kind used in the sensing chain of grid-connected
converters, \sys{} ran the experiment in 2\,h 39\,min with no human action, and the board
removed 50.6\,dB at 60\,Hz. It was the first run of the experiment: the agent resolved by itself the faults it
met on the way, and from the data and its own run it proposed improvements to the design and to the experiment. Rehearsal exposed five defects before any material was used, including a test
plan that would have rejected a correct board. A live display shows the bench at work, and a thinking screen shows what
the agent thinks. To our knowledge, \sys{} is the first system in power electronics in which an agent closes the
physical design--build--test--learn loop for circuit boards.
\end{abstract}


\section{Introduction}
\label{sec:intro}

\begin{figure*}[t]
\centering
\includegraphics[width=\textwidth]{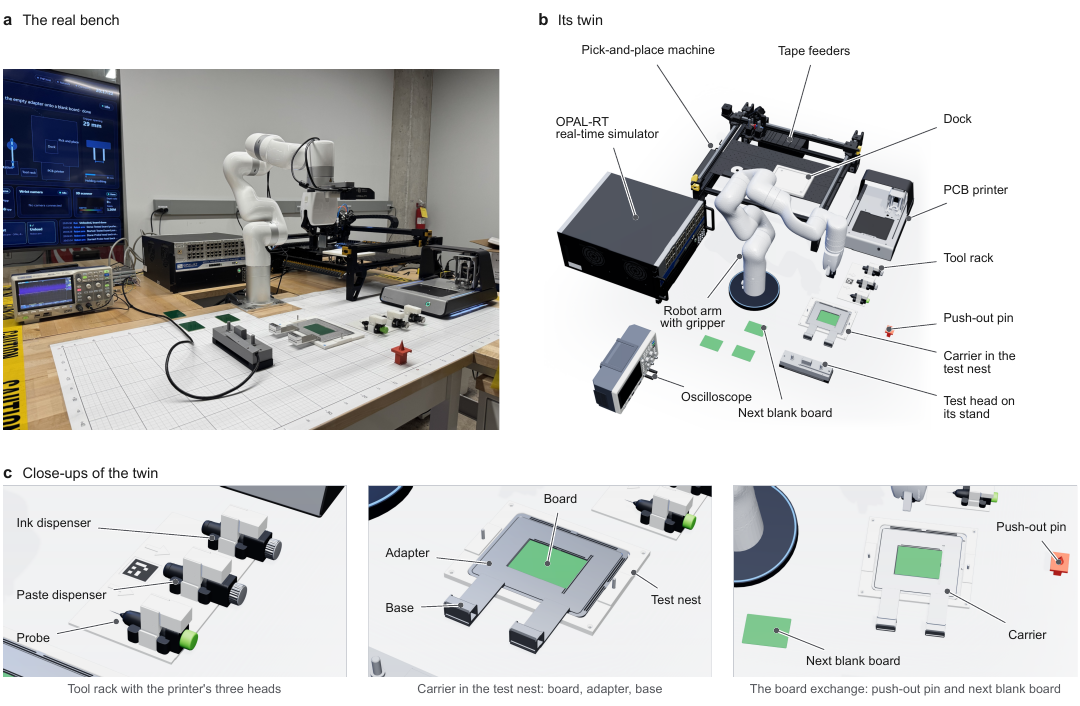}
\caption{\textbf{The \sys{} bench and the agent's twin of it.} (a)~The physical bench. (b)~The twin of the same bench at the start of a run. Every machine and fixture is placed where the
bench scan found it: the robot arm with its gripper, the printer, the pick-and-place machine with its dock and tape
feeders, the tool rack, the test nest holding the carrier, the test head on its stand, the push-out pin, the next
blank board, the simulator and the oscilloscope. (c)~Close-ups of the twin. Left: the tool
rack holds the printer's probe, paste dispenser and ink dispenser. Middle: in the carrier, the board sits in an adapter
and the adapter sits in a base. Right: at the end of a run, the tested board is pushed out of the adapter on the
push-out pin, and the empty adapter is set onto the next blank board.}
\label{fig:overview}
\end{figure*}

Power electronics sits between every energy source and every load. It feeds solar and wind power into the grid,
charges and drives electric vehicles, and supplies data centers. Its hardware improves through a loop. Engineers state
what a board must do, design it, build it, measure it and learn from the gap; then they go around again. Before a
board or a controller meets a real grid, the field verifies it \emph{hardware-in-the-loop} (HIL). The real board is
connected to a real-time simulator, which produces the board's input signals and records its response in real
time~\cite{faruque2015realtime}. Every turn of this loop still passes through a human engineer. The engineer reads the
requirement, draws the schematic, lays out the board, carries it from machine to machine, wires it to the simulator
and interprets the result.

Language models have started to help with the design step. In power electronics, they choose modulation strategies and
converter parameters~\cite{lin2025pegpt,liu2026phia}, draft converter designs~\cite{liu2026ietpe} and generate
schematics and board layouts from a requirement~\cite{sahu2026powercircuit}. In other fields, large language model
(LLM) agents fix real software issues~\cite{jimenez2024swebench,yang2024sweagent}, write hardware description
code~\cite{chang2023chipgpt,liu2023verilogeval,thakur2023autochip} and design analog circuits~\cite{lai2025analogcoder}.
Robots guided by language grasp, place and assemble objects~\cite{ahn2022saycan,liang2023code,huang2023voxposer,
brohan2023rt2}. The physical part of the hardware loop, however, is still done by people. Each turn of the loop crosses
several machines, materials and instruments, and no autonomous system has handled all of them together.

\emph{Self-driving laboratories}, in which robots run experiments that software chooses, show what closing such a loop
is worth. Guided by optimizers or by LLM agents, they have searched photocatalyst mixtures~\cite{burger2020mobile},
synthesized computationally predicted inorganic compounds~\cite{szymanski2023alab}, and planned and run chemical
reactions~\cite{boiko2023coscientist,bran2024chemcrow}. They compress months of bench work into
days~\cite{abolhasani2023rise,tom2024sdl}. Most of these systems vary the \emph{recipe} of a fixed workflow: the steps
stay the same and only the quantities change. Even an agent that builds its own chemistry laboratory produces samples,
not devices that must first be designed~\cite{bai2026selfbuilding}. Hardware asks for more, because a new circuit is a
new object. It changes the pattern to print, the paste to dispense, the parts to place, the way the board is contacted
and the test that judges it. Language-model tools for electronic and power-electronic design, on the other hand, stop at
simulation or the layout, and people still build and measure the board. What is missing is an agent that is engineer and
technician at once: one that designs a board, builds it on real machines, measures it, understands the result and
decides what to change.

We present \sys{}, an embodied agent that closes the design--build--test--learn loop for power-electronics hardware
(Fig.~1). \figref{fig:overview} shows its bench, next to the twin that the agent built of it. \sys{} works in
four stages, and each stage builds on the one before.
\begin{enumerate}
  \item \textbf{It knows its machines.} For every machine, the agent keeps a record of what the machine is, how to
  connect to it, what its limits are, and what has worked or failed on it. Every fact in the record states its source.
  \item \textbf{It builds its own twin of the bench.} A 3D scanner on the robot arm scans the bench from several directions.
  The agent merges the views into a panorama of the whole bench and places the complete CAD model of every machine and
  fixture (a holder for boards or tools) where the scan found it. The result is a \emph{digital twin} of the bench.
  The twin is \emph{executable}: the programs that control and judge the real bench also run inside it.
  \item \textbf{It rehearses the experiment in the twin.} From a need stated in plain language, the agent sets the
  requirement, designs the circuit and the printed circuit board (PCB), and turns the design into programs for every
  machine. It
  then runs the whole build and test in the twin. The rehearsal covers the physical side of the experiment: the motion
  of the arm, the heating of the board and its holder, the time of every step and the response of the circuit. The
  agent fixes every defect it finds there.
  \item \textbf{It runs the experiment on the bench and learns from it.} A six-axis arm, a desktop PCB printer and a
  pick-and-place machine build the board. An OPAL-RT real-time simulator tests it hardware-in-the-loop. The agent then analyzes the
  measured response, explains each departure from the design, and proposes the next design, which it draws and
  rehearses.
\end{enumerate}

Four principles make these stages work on a real bench. First, the agent operates machines as a skilled person does:
no machine's hardware or firmware is changed. The arm changes the printer's tool heads, and the agent operates the
printer's own application as a person would click through it. Second, the agent checks its own work. Its 3D scanner
and the machines' own cameras and sensors confirm every step, and no person inspects anything during a run. Third, every
physical action is accountable. Before each action, the agent writes its goal, plan, expected result and stopping
condition into an attempt ledger. After the action, it saves the full output as evidence and records what it learned.
A \emph{session} is one continuous run of the agent, and the language model remembers nothing from earlier sessions.
The records therefore let any later session resume the work. Fourth, the work is visible. During a run, a live display
shows the bench and the data of every machine, and a thinking screen shows what the agent is thinking, in plain words.

Our benchmark is a 60\,Hz notch filter. A grid-connected converter must measure small quantities next to the large
60\,Hz fundamental of the grid, such as harmonics or a DC offset in its current. Its low-level sensor signals also pick
up 60\,Hz hum. A 60\,Hz notch filter removes the fundamental or the hum, so that the small quantity can be measured. The
requirement of such a filter is precise and its test is exact, so the behavior of the loop can be judged without doubt.
A person stated the need in plain words: a notch filter at signal level, as hardware. The agent set the requirement
itself: remove 60\,Hz mains hum from a $\pm$10\,V signal by at least 40\,dB, with a flat passband. It designed the
circuit and the board in KiCad, and it ran the rehearsal of their build in the twin. The person read the agent's plan and agreed
to it in one sentence typed on a screen, which started the run. In \sys{}, a person states the need and agrees to the
agent's plan; the agreement is a safety step of this first stage, and once the system is mature, the need alone will
start the work. \sys{} then ran it to the end with no human
action, in 2\,h 39\,min: it surveyed the bench, built the board, tested it on the simulator, and pushed the
tested board out of its holder to load the next blank one. The board's notch lay at 60.3\,Hz, and it removed
50.6\,dB at 60\,Hz. From the measured response, the agent explained why the passband sagged at low frequency: the
filter network's output impedance met the load on its output, and it proposed the next board, which it drew in KiCad and rehearsed in the twin. This was the first run of the
experiment. On the way, the agent met faults of the kinds common in computer and engineering practice and
resolved each by itself, and at the end it proposed improvements to both the board and the experiment, which a platform
that executes a fixed recipe does not do. Rehearsal in the twin
had caught defects before they could waste a board or cause a wrong verdict. In one, the planned frequency sweep would
have failed a correct board. In another, the pattern that the printer draws before each print to start the ink flowing
would have crossed the circuit.

\noindent Our contributions are:
\begin{itemize}
  \item \textbf{Autonomous Hardware Loop for Power Electronics.} To our knowledge, \sys{} is the first system in power
  electronics, and for circuit boards in general, in which an agent closes the complete design--build--test--learn
  loop. From a need stated in plain language, it sets the requirement, designs the circuit and the board, builds the
  board on unmodified desktop machines, verifies it hardware-in-the-loop on the simulator, analyzes the measurement and turns it into the
  next design.
  \item \textbf{An Executable Digital Twin Built by Scanning, for Physical Rehearsal.} The agent scans its bench, merges
  the views into a panorama and builds the twin on its own. The twin runs the bench's own motion planners,
  safety checks and physical models, so every multi-machine build is rehearsed from start to end, including its motion,
  heat, time and circuit response, before it runs.
  \item \textbf{Embodied Machine Operation.} One arm and one agent operate the printer, the pick-and-place machine and the
  simulator without changing their hardware or firmware. The arm changes the printer's tool heads, a task the
  printer was never designed for, and exchanges boards by itself. The agent adapts the build to each board: it reasons
  about the board's surface at every stage, has it mapped so that the nozzle follows it, and decides how much
  unevenness each stage may have.
  \item \textbf{Self-Checking, Accountable and Visible Autonomy.} The agent confirms every step with its 3D scanner
  and the machines' own cameras and sensors, and measures the board with two instruments. An attempt ledger with a source label
  on every fact makes every physical action auditable and resumable across sessions. Two screens show people what the
  bench is doing and what the agent is thinking.
\end{itemize}

\section{Related Work}
\label{sec:related}

\begin{table*}[t]
\centering
\caption{Capabilities of \sys{} and of representative systems from the five lines of work of \secref{sec:related}:
self-driving laboratories, LLM agents for science and for robots, language models for electronic design, language
models for power electronics, and digital twins and real-to-sim. Physical execution: the system acts on real equipment. Builds a device: it produces a
working device, not a sample or a design. Design from requirement: it creates the design from a stated goal. Closed
loop: it chooses the next experiment or design from measured results, without a person. Multi-machine workflow: one
task uses several machines. Self-surveyed twin: it builds its own digital twin of the workspace from its sensors.
Rehearsal before execution: it checks the planned actions in simulation before acting. \cmark: full, \pmark: partial, \xmark: absent; marks as reported in the cited papers.}
\label{tab:capabilities}
\setlength{\tabcolsep}{3.4pt}
\begin{tabular}{@{}llccccccc@{}}
\toprule
System & Domain & \makecell{Physical\\execution} & \makecell{Builds a\\device} & \makecell{Design from\\requirement} &
\makecell{Closed\\loop} & \makecell{Multi-machine\\workflow} & \makecell{Self-surveyed\\twin} &
\makecell{Rehearsal before\\execution} \\
\midrule
Mobile robotic chemist~\cite{burger2020mobile} & Photocatalysis & \cmark & \xmark & \xmark & \cmark & \cmark & \xmark & \xmark \\
A-Lab~\cite{szymanski2023alab} & Inorganic materials & \cmark & \xmark & \pmark & \cmark & \cmark & \xmark & \xmark \\
Coscientist~\cite{boiko2023coscientist} & Chemistry & \cmark & \xmark & \cmark & \pmark & \pmark & \xmark & \xmark \\
ChemCrow~\cite{bran2024chemcrow} & Chemistry & \pmark & \xmark & \cmark & \xmark & \xmark & \xmark & \pmark \\
AnalogCoder~\cite{lai2025analogcoder} & Analog circuits & \xmark & \xmark & \cmark & \pmark & \xmark & \xmark & \xmark \\
PE-GPT~\cite{lin2025pegpt} & Power electronics & \xmark & \xmark & \cmark & \pmark & \xmark & \xmark & \xmark \\
Power Circuit AI~\cite{sahu2026powercircuit} & Power electronics & \xmark & \xmark & \cmark & \xmark & \xmark & \xmark & \xmark \\
RialTo~\cite{torne2024rialto} & Manipulation & \cmark & \xmark & \xmark & \xmark & \xmark & \pmark & \pmark \\
OptoMate~\cite{uddin2026optics} & Free-space optics & \cmark & \cmark & \cmark & \pmark & \pmark & \xmark & \pmark \\
Self-Building Labs~\cite{bai2026selfbuilding} & Lab automation & \cmark & \xmark & \cmark & \pmark & \cmark & \pmark & \cmark \\
\midrule
\textbf{\sys{} (ours)} & \textbf{Electronic hardware} & \cmark & \cmark & \cmark & \cmark & \cmark & \cmark & \cmark \\
\bottomrule
\end{tabular}
\end{table*}

\tabref{tab:capabilities} compares \sys{} with ten representative systems from five lines of work. Only \sys{} has
every capability.

\para{Self-driving laboratories.}
Automating the scientific loop goes back to robot scientists that formed and tested hypotheses in
genomics~\cite{king2009automation}. Self-driving laboratories now combine robotic execution with Bayesian optimization
or active learning in chemistry and materials science~\cite{hase2019next,macleod2020selfdriving,shields2021bayesian,
abolhasani2023rise,tom2024sdl}. The \emph{mobile robotic chemist} moves between stations to optimize a photocatalyst
mixture~\cite{burger2020mobile}. Other mobile robots share standard instruments for exploratory
synthesis~\cite{dai2024mobile}. \emph{A-Lab} synthesizes computationally predicted inorganic
compounds~\cite{szymanski2023alab}, and a closed-loop platform makes and characterizes perovskite solar
cells~\cite{gao2026perovskite}. In these systems the workflow is fixed in advance, and the optimizer chooses the
workflow's parameters. In \sys{}, each design is a different object, so the programs that build it change from one
iteration to the next. The agent writes these programs itself.

\para{LLM agents for science and for robots.}
LLM agents plan and run chemical experiments through liquid handlers and cloud laboratories, as in
\emph{Coscientist}~\cite{boiko2023coscientist} and \emph{ChemCrow}~\cite{bran2024chemcrow}, and in related
work~\cite{yoshikawa2023chemistry}. They also carry out the computational research cycle from idea to
paper~\cite{lu2024aiscientist}. Software agents reason and act through tools~\cite{yao2023react,wang2023voyager,
yang2024sweagent}. Robot policies that follow language instructions link language to physical
action~\cite{ahn2022saycan,huang2022inner,liang2023code,huang2023voxposer,brohan2023rt2}. \emph{OptoMate} uses a
language model and a robot to assemble free-space optical setups from a stated goal~\cite{uddin2026optics}. Another
robot redesigns assembled block structures after checking them~\cite{khendry2026idfra}. In a quantum experiment,
control code written by an agent runs only after it passes checks against per-device limits in a simulation
of the hardware~\cite{wang2026artiq}. \sys{} brings these lines together for a whole bench and for working electronic
hardware. The agent writes and runs the programs of several machines. It checks every program in a twin of the bench
before running it, and it judges each build by measuring it.

\para{Language models for electronic design.}
LLMs generate and repair register-transfer-level digital designs~\cite{chang2023chipgpt,liu2023verilogeval,
blocklove2023chipchat,thakur2023autochip}. \emph{AnalogCoder} designs analog circuits~\cite{lai2025analogcoder}. These
systems close their loops through circuit simulation and testbenches. \sys{} closes the loop through fabrication and
measurement: its feedback is the measured frequency response of a board that it printed, assembled and tested itself.

\para{Language models for power electronics.}
Language models have also reached the design desk of power electronics. \emph{PE-GPT} combines an LLM with
physics-informed neural networks to design converter modulation and circuit parameters~\cite{lin2025pegpt}. Autonomous
LLM agents produce explainable modulation designs~\cite{liu2026phia}. Customized LLMs generate and verify converter
designs~\cite{liu2026ietpe}. \emph{Power Circuit AI} uses several cooperating agents to turn a requirement into a
converter's schematic and PCB layout~\cite{sahu2026powercircuit}. The outputs of these systems are designs. They are
validated in simulation, or on prototypes that engineers build by hand. \sys{} takes the next step. The agent builds the
board itself and tests the real board on the kind of real-time simulator that power-electronics engineers use for HIL
testing~\cite{faruque2015realtime}. The loop is therefore closed by measurements of the real board, not by a simulation
of it.

\para{Digital twins and real-to-sim.}
Digital twins mirror physical systems in order to predict and prevent unwanted behavior~\cite{grieves2017digital,
tao2019digital}. In robotics, real-to-sim methods rebuild a scene from scans and train robot policies in it that
transfer to the real world. In \emph{RialTo}, a person scans the scene with a phone and adds the moving
parts~\cite{torne2024rialto}. Neural rendering such as Gaussian splatting~\cite{kerbl2023gaussian} makes simulated
laboratories look photorealistic~\cite{dexbiolab2026}. In \emph{Self-Building Labs}, an agent that builds its own
chemistry laboratory assembles a twin from photographs, documents and vendor models, and checks its protocols there
before running them~\cite{bai2026selfbuilding}. The twin of \sys{} differs in two ways. First, the agent builds it by
itself, by scanning the bench with a 3D scanner on the arm, and it scans again before every run to confirm that the
bench still matches it. Second, it is executable: it runs the bench's own planners, checks and physical models. The
agent can therefore rehearse a whole multi-machine build, from the motion of the arm and the heating of the board to the
predicted measurement, before acting.

\section{The \sys{} System}
\label{sec:system}

\begin{figure*}[t]
\centering
\includegraphics[width=\textwidth]{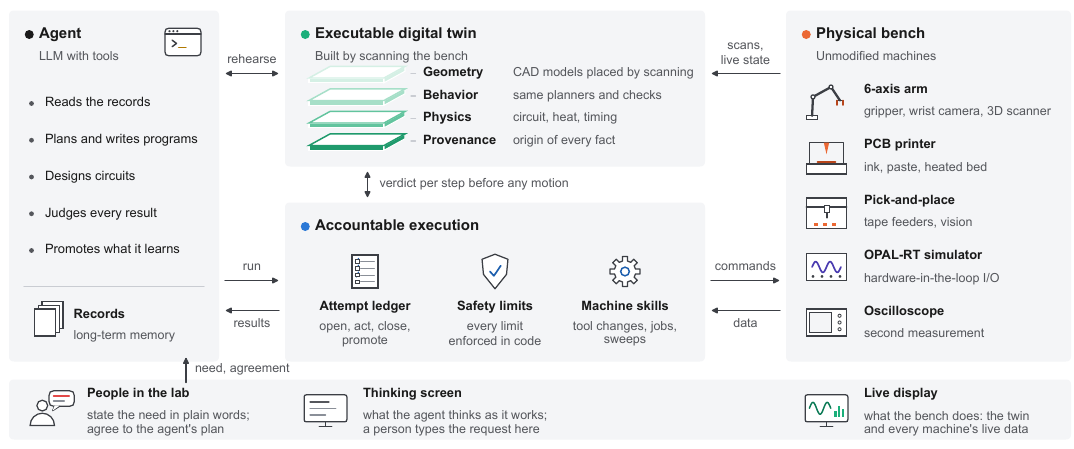}
\caption{\textbf{Architecture of \sys{}.} The agent reads and writes the records, its long-term memory, and acts
through two channels. In the twin, the agent asks questions and rehearses. The twin's four layers
(geometry, behavior, physics, provenance) mirror the bench. Through the accountable execution layer, the agent runs
programs on the real machines. This layer logs every attempt, enforces every safety limit in code and holds the
machine skills. The bench consists of unmodified machines. Scans and the machines' live states keep the twin in step
with the bench. Two screens show people what the bench does and what the agent thinks. A person states the need in
plain language and agrees to the agent's plan; once the run starts, no person intervenes.}
\label{fig:architecture}
\end{figure*}

\sys{} has four parts (\figref{fig:architecture}).
\begin{itemize}
  \item The \textbf{agent} reasons, writes code and decides what to do next. It keeps \emph{records}: text files
  about every machine and every task, which serve as its long-term memory.
  \item The \textbf{executable digital twin}, the \emph{twin} for short, holds a 3D model of every machine and
  fixture, and models that predict how the bench will move, heat up and measure. The agent uses it to try out programs
  before it runs them on the real machines.
  \item The \textbf{accountable execution layer} (the execution layer, for short) runs every program on the real
  machines. It holds the \emph{machine skills}, the programs that operate each machine (tool changes, flows of the
  printer's application, placement jobs, test sweeps). It enforces the safety limits and logs every action.
  \item The \textbf{physical bench} is the set of unmodified machines that build and test the boards.
\end{itemize}
The agent works in four stages. It learns what its machines are. It scans the bench and builds the twin. It sets the
requirement from a need, designs a board for it and rehearses the whole experiment in the twin. Then it runs the
experiment on the bench, measures the board and improves the design. This section describes the bench and the agent, how the agent knows, operates and
checks its machines, and how it keeps every action accountable and safe.

\subsection{The Physical Bench}
\label{sec:bench}

The bench is built from standard equipment, chosen for what it does and used as delivered.
\begin{itemize}
  \item A \textbf{six-axis collaborative robot arm} with a two-finger gripper. A camera and a 3D scanner are mounted on
  its wrist. The arm carries every board and every tool on the bench.
  \item A \textbf{desktop PCB printer}. It deposits conductive silver ink and solder paste from interchangeable tool
  heads. Another tool head, the probe, touches the board to measure its height. The printer's heated bed cures the ink and
  reflows (melts) the solder paste.
  \item A \textbf{pick-and-place machine} with tape feeders and two cameras, one looking down at the board and one
  looking up at the part on the nozzle. A small embedded computer runs its control software, and the agent controls
  the embedded computer.
  \item An \textbf{OPAL-RT real-time simulator} with analog outputs and inputs. It generates the test signal and records
  the board's response in real time; the agent builds, loads and runs its test models.
  \item An \textbf{oscilloscope}, which measures the board's input and output as a second instrument, with its own
  acquisition; the agent sets it up and reads it.
  \item Two screens for the people in the lab: a large \textbf{live display}, which shows the twin and the live data of
  every machine, and a small \textbf{thinking screen}, which shows the agent's reasoning and takes requests typed in
  plain language.
\end{itemize}

\begin{figure*}[t]
\centering
\includegraphics[width=\textwidth]{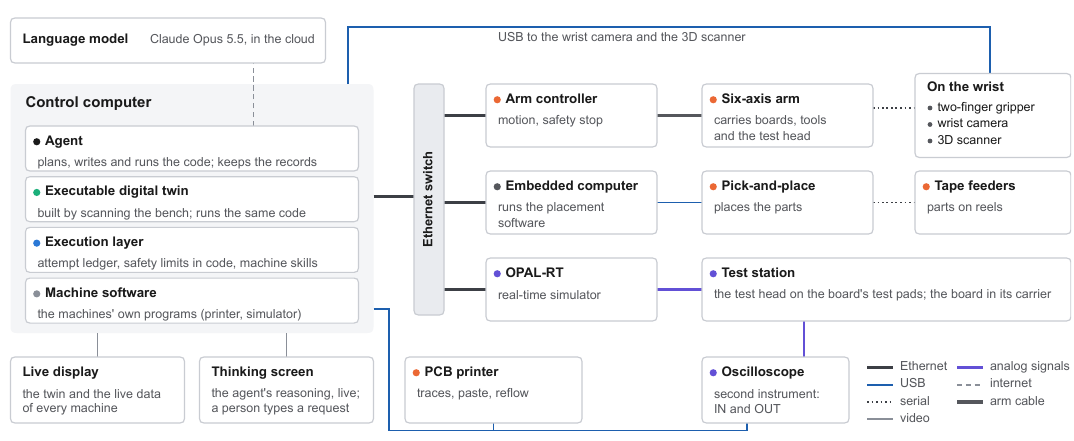}
\caption{\textbf{How the bench is connected.} Machines are shown by their role. Line styles show the type of each
link. The agent and its twin run on the control computer; the agent's reasoning comes from a language model in the
cloud.}
\label{fig:bench}
\end{figure*}

\figref{fig:bench} shows how the parts are connected. The agent's program, the twin, the execution layer and the
machines' own software run on a control computer. The agent's reasoning comes from a large language model, which runs
in the cloud and is reached over the internet. Through an Ethernet switch, the control computer reaches the arm's
controller, the simulator and the embedded computer of the pick-and-place machine. The printer, the
oscilloscope, the wrist camera and the 3D scanner connect to the control computer by USB. The simulator's analog
outputs and inputs reach the board through shielded cables and a test head that touches the board. The two screens are
driven by the control computer.

\subsection{How a Board Moves: the Carrier}
\label{sec:carrier}

\begin{figure*}[t]
\centering
\includegraphics[width=\textwidth]{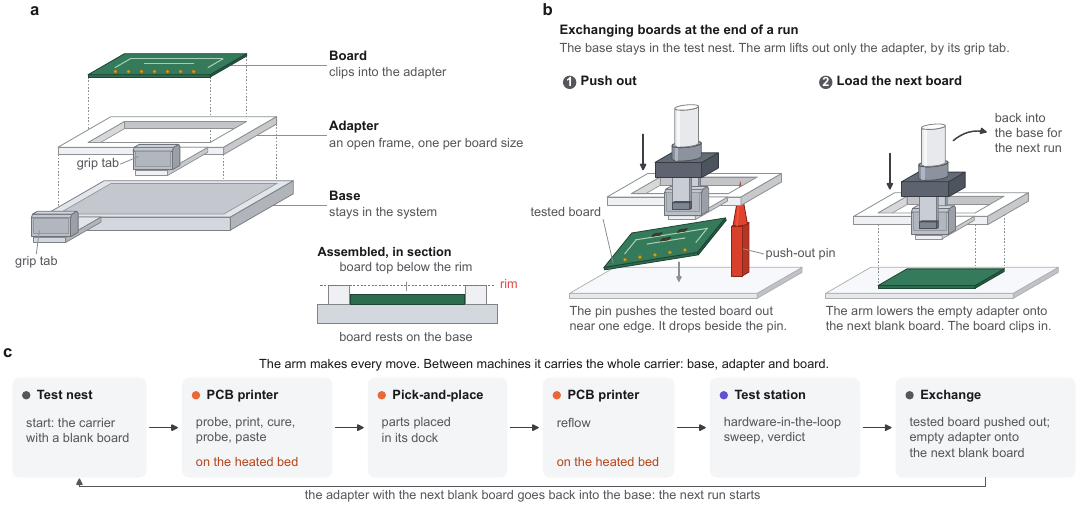}
\caption{\textbf{How a board travels through the bench, and how boards are exchanged.} (a)~The board clips into an
adapter, an open frame. The adapter sits in a base. Through the opening, the board rests directly on the base, and its
top stays below the adapter's rim. The adapter and the base each have a grip tab, by which the arm holds them. A board
of another size uses another adapter. (b)~Board exchange at the end of a run. The base stays in the test nest, and the
arm moves only the adapter. (1)~The arm lowers the adapter over a push-out pin, which pushes the tested board out near
one edge; the board drops onto the table. (2)~The arm lowers the empty adapter onto the next blank board, which clips
in. (c)~The route of one board. The arm carries the whole carrier from machine to machine, and the whole carrier is
heated on the printer's bed. After the test, the exchange loads the next blank board, and the next run starts again
in the test nest. Schematic, not to scale.}
\label{fig:carrier}
\end{figure*}

Each board moves through the bench in a two-part \emph{carrier} (\figref{fig:carrier}a). The board clips into an
\emph{adapter}, an open frame made for one board size. The adapter sits in a \emph{base}. Through the open frame, the
board rests directly on the base, which passes the heat of the printer's bed to the board. The base and the adapter
each have a grip tab, so the arm can pick up the whole carrier or the adapter alone.

The arm carries the whole carrier from machine to machine (\figref{fig:carrier}c): to the printer's bed, to a dock on
the pick-and-place machine and to the test station. The carrier stays together through every step, including the
heating on the printer's bed. At the test station, the carrier sits in a holder that we call the \emph{test nest}, and
a \emph{test head} touches a row of test pads on the board.

Boards are exchanged by moving the adapter alone (\figref{fig:carrier}b). After the test, the arm lifts the adapter,
with the tested board, out of the base. It lowers the adapter over a \emph{push-out pin} that stands on the table. The
pin reaches the board from below, through the open frame, and pushes it out; the board drops onto the table. The arm
then sets the empty adapter onto the next blank board, which waits on the table, and the board clips into it. The
adapter with the new board returns to the base for the next run. The base never leaves the system, and a board of
another size only needs another adapter.

\subsection{The Agent and Its Knowledge of the Machines}
\label{sec:agent}

The agent's reasoning comes from \llmname{}, a large language model. The agent works in a code repository that holds
the bench's control code, the bench's settings and the records. It reads and writes files, runs programs and reads
their output. The agent wrote the code that drives the bench: device drivers, machine skills, the twin, the programs
that turn a design into machine files, the test programs for the simulator and the two screens. It uses, but did not
write, the machines' own software and the vendors' programming libraries.

The agent knows each machine through that machine's records.
\begin{itemize}
  \item \emph{Facts:} what the machine is, how it is connected, its settings, and its limits: speeds,
  temperatures, working space, emergency stop and safe pose.
  \item \emph{Proven procedures:} step-by-step sequences that the agent has seen work on the bench.
  \item \emph{Known pitfalls:} each failure that the agent has met, with its symptom, its cause and its fix.
  \item \emph{A log} of every action on the machine, and the machine's \emph{manuals} and \emph{CAD model}.
\end{itemize}
Before the agent acts on a machine, it reads these records. After acting, it writes back what it learned. The language
model keeps no memory from one session to the next, and work often continues in a new session. The records are
therefore the agent's long-term memory, and any later session can continue the work from them.

When a new machine joins the bench, the agent creates its records and fills in what the manuals state. Its first
actions on the machine only read: identify the machine, connect to it and read its status. The machine may not move
under the agent's control until its emergency stop, limits and safe pose are recorded from a source other than an
assumption.

We can also replace \llmname{} with another language model, as our conditions and needs require: the records and the
bench's code do not depend on the model.

\subsection{Operating Unmodified Machines}
\label{sec:skills}

A central design choice of \sys{} is that no machine's hardware or firmware is changed. The agent uses each machine
through the interfaces that come with it: its own application, its command interface or its programming library. The
bench therefore stays reproducible with standard equipment, and a new machine can join without being rebuilt.

\para{Robotic tool changes.}
The printer has three tool heads: the probe, the ink dispenser and the paste dispenser. For one board, a head is
mounted four times: the probe, the ink dispenser, the probe again and the paste dispenser. Each mounting and each
removal is one tool change, so one board needs eight tool changes. The printer expects a person to change the heads
by hand; in \sys{} the arm does it. The agent first sends the printer's carriage, which holds the tool head, to a fixed
tool-change position. The \emph{tool rack} is located by a printed marker~\cite{garrido2014aruco} on it, which the 3D
scanner on the wrist has found. The arm takes a head from its slot and seats it on the carriage, in the order the manufacturer prescribes for
a person. Removing one head and mounting the next takes about three minutes.

\para{The printer's own application.}
The printer's application plans the print paths, runs the printer's calibration and controls the heater. The agent
operates it as a person would: it reads each screen and presses its buttons by their labels. A fixed sequence of
screens, such as a print, is called a \emph{flow}. After every press, the agent compares the screen with the one it
expects, and it stops on any error message or timeout. The agent heats the bed itself, with a heating program that
holds the bed at the temperature and for the time that the run's settings give. The agent also goes beyond
operating the application, without changing it. As installed, the application refuses to print on a board that is
not flat enough. The agent adapts the application to the board at each stage of the build, so the printer works on
real, uneven boards, while every other check of the application stays in force.

\para{The agent adapts the build to the board.}
A board is never perfectly flat, and a nozzle that passes a fraction of a millimetre too high or too low draws a poor
trace. The agent therefore treats the shape of each board as something to reason about, not as a fixed setting. Before
each deposit, it has the probe touch the board at many points. This gives a height map of the surface, and the nozzle
follows the map, so that it keeps the same small gap above every part of the board, even where the board bends. How
much unevenness is acceptable depends on the stage, and the agent works it out from what it knows about the board at
that moment. On a bare board, the map covers bends and unevenness, but neighbouring points must not differ by much:
a larger step means that the board is not seated in its adapter, and the run stops before any ink is spent.
After printing, the traces and pads stand above the bare surface, so the probe will see larger differences, and they are
expected. Before the paste, the agent therefore relaxes the limit so that it covers the height of the printed layer.
After the reflow, it sets the strict limit again for the next blank board. No person sets these values: the agent
chooses them for each stage and applies them between the printer's steps, and every height map is judged against
them.

\para{The pick-and-place machine through its own computer.}
The pick-and-place machine is run by its open-source control software, on a small embedded computer beside it, which
the control computer reaches over the network. The agent does not change this software either. It wrote a small
program of its own, which the software loads at start-up through its own scripting hook. The program takes the agent's
commands over the network, such as load a job, home, start, stop or park the head, and carries out each one by pressing
the software's own buttons, so it refuses whatever the software's interface would refuse. It also reports the
machine's state, whether a job is running, where the head is and what each nozzle holds, and the events of each job:
its start, its end or an error. The agent generates the placement job from the board's design files and copies it to
the embedded computer. Before the arm reaches into the machine, the program must report that entry is safe: no job is
running, the placement head is parked, and the nozzles are empty and raised. We call this check the
\emph{interlock}.

\para{The simulator and its test models.}
The agent controls the simulator from the control computer, through the simulator's own programming library. For each
test, it writes a \emph{test model}, a program that plays the test signal and records the board's response, and its
own programs build the model, generate its code and compile it on the simulator. Before a sweep, the agent's program
confirms that the simulator holds the expected test model, ready to load, with every link between the model's signals
and the analog channels in place, and after loading that every signal it needs is there; otherwise the sweep does not
start. It then runs the model in real time, sets the frequency and the amplitude of the test signal point by point and
reads the response. Whatever happens, the sweep ends with the output at zero and the model stopped and unloaded.

\para{The oscilloscope.}
The agent controls the oscilloscope over USB, with the oscilloscope's own commands. Before it measures, it reads back
the probe attenuation and the coupling of both channels, and it refuses to measure if they differ from its records.
During the sweep, it sets the time base and the volts per division for each frequency, takes a capture again when a
channel is clipped or too small, and changes nothing but the oscilloscope's acquisition settings.

\subsection{The Agent Checks Its Own Work}
\label{sec:selfcheck}

No person inspects any step of a run. The agent confirms each step with its own sensors and with the machines' own
reports.
\begin{itemize}
  \item \emph{The bench.} Before a run, the 3D scanner surveys every machine and fixture that the run uses. The agent
  compares each one with its place in the twin. A place that does not match stops the run.
  \item \emph{The arm.} The 3D scanner locates the tool rack by its printed marker. The gripper's
  opening after each grip tells whether the arm holds the expected part.
  \item \emph{The printer.} The printer reports its carriage position and the tool head it carries. The probe measures
  the board's surface before each deposit. The bed's temperature is read throughout each heating, and the arm grips
  the carrier only once the bed is cool enough to touch.
  \item \emph{The pick-and-place machine.} The down-looking camera finds the board's printed reference marks
  (\emph{fiducials}) and so measures where the board lies. The up-looking camera checks each part on the nozzle before
  it is placed. The interlock confirms that the machine is at rest before the arm enters, and the placement job counts as done
  only when the machine's software reports that it finished; an error stops the run.
  \item \emph{The simulator and the oscilloscope.} Before a sweep, the agent checks the simulator's test model and its
  links to the analog channels, and the oscilloscope's probe and channel settings. The recorded input confirms that the
  test head touches the board, and a sweep in which the simulator missed any real-time step is flagged as suspect. The
  oscilloscope measures the same signals as the simulator, as a second reading with its own acquisition.
\end{itemize}
A failed check stops the run at the last good step. The run saves its progress, so it can resume from that step, even
in a later session of the agent.

\subsection{Accountable Execution}
\label{sec:ledger-method}

Every action whose result is not known in advance is an \emph{attempt}. Examples are a command sent to a machine, a
motion of the arm, and a test to find out how something works. Each attempt has four steps.
\begin{enumerate}
  \item \emph{Open.} Before acting, the agent writes the goal, the exact plan, the expected result, the condition for
  stopping and the state of the bench. The execution layer refuses to run the attempt until all of these are written.
  \item \emph{Act.} The execution layer runs the plan and saves the complete output as evidence.
  \item \emph{Close.} The agent records the outcome: success, failure, partial, aborted or unverified.
  \item \emph{Promote.} The agent moves what it learned into the machine's records (facts, proven procedures or known
  pitfalls), and each lesson cites the attempt it came from.
\end{enumerate}
Together, the attempts form the \emph{attempt ledger}. Larger choices, such as how to build a fixture, are kept as
\emph{design decisions}, each with the options that were rejected. The records keep observations apart from
interpretations. Every fact also carries a label that states its origin: for example a measurement in a named
attempt or a page of a vendor manual. We call this label the \emph{provenance} of the fact.

\subsection{Safety by Construction}
\label{sec:safety}

Every quantity that matters for safety has a pass/fail limit. The limit is enforced in code and covered by software
tests. The limited quantities include the arm's joint and Cartesian speeds, the bed's temperature and heating time, and
the \emph{keep-out boxes} around every machine, which are regions the arm must not enter. The keep-out boxes follow the
machines' reported state. For example, the arm may enter the area of the printer's gantry, the bridge that carries the
carriage, only while the printer reports its carriage at the tool-change position. A program may move or heat a
machine only when it is given explicit permission. The heating program switches the heater off if the bed exceeds its
limit, if its reading stops, or if the bed takes too long to warm. For the test, the simulator's output is held within
$\pm$2\,V by a limiter inside the test model. The sweep stops if any gain exceeds +3\,dB: the notch board amplifies
nothing, so a higher gain means a fault in the board or the wiring. No motion may depend on a value whose provenance is
an assumption.

People stay safe as well as machines. Every new motion runs first at reduced speed, with a person at the emergency stop.
During every run, as the laboratory's safety rules require, a person stands by the arm's emergency stop. This person
watches but does not touch the bench.

\section{Building the Executable Digital Twin}
\label{sec:twin}

\begin{figure*}[t]
\centering
\includegraphics[width=\textwidth]{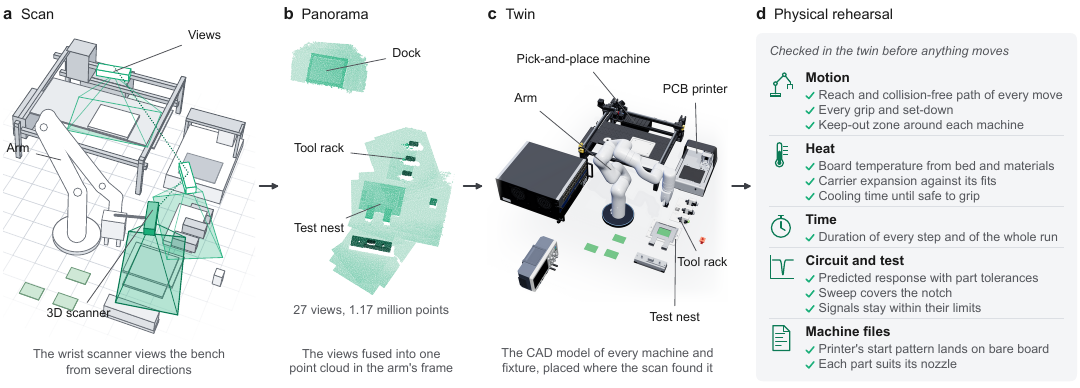}
\caption{\textbf{From scans to the twin, and a rehearsal in it.} (a)~The 3D scanner on the arm's wrist views the
bench from several directions; each cone is one view. (b)~The agent fuses the views into one panorama of the bench, a
point cloud in the arm's frame, coloured by height above the surface below. (c)~The agent places the complete CAD model of every machine,
fixture and tool where the scan found it. The result is the executable digital twin.
(d)~Before anything moves, the agent rehearses the experiment in the twin. It checks motion, heat, time, the
circuit and its test, and the machine files.}
\label{fig:twin}
\end{figure*}

The agent's second stage is to build its own twin of the bench. A digital twin mirrors a physical system so that its
behavior can be predicted before it happens~\cite{grieves2017digital,tao2019digital}. The twin of \sys{} has two
further properties. It is \emph{self-built}: the agent builds it from its own scans and keeps it up to date by scanning
again. And it is \emph{executable}: the code that drives the bench runs inside it unchanged. A prediction in the twin
therefore comes from the very program that will drive the bench, executed on a model of the bench instead of the bench
itself.

\subsection{From Scans to the Twin}
\label{sec:survey}

The 3D scanner rides on the arm's wrist. To survey the bench, the agent plans views of each machine and fixture from
several directions. It checks the path of the arm in the twin, moves the arm and takes the scans. Because the agent
knows where the arm is at every moment, it merges all views into one panorama of the bench (\figref{fig:twin}a,b).

From the arm, the scanner sees only part of each object: mostly its top faces, seen from a few directions. Sides,
undersides and hidden corners are missing. The agent therefore does not build the scene from the scans alone. It knows
what each machine and fixture is, and it has the complete CAD model of each. It places each model into the panorama
where the scan matches it best, so the twin holds every object whole, at its measured place (\figref{fig:twin}c).
The scene of the twin is built in this way without a modeling team and without a person placing anything.

The twin stays current. Before every run, the agent scans the bench again and confirms each object at its place in the
twin. During the run, every program that moves something, for example a carrier set down in the dock or a tool head
taken from the rack, updates the twin at once.

\subsection{Four Layers of the Twin}
\label{sec:layers}

The twin has four layers (\figref{fig:architecture}).

\para{Geometry.}
Every machine, fixture, tool and carrier appears as its CAD model, placed in the frame of the arm. Each model comes
from the machine's maker or from the agent, which designed the fixtures itself. Each placement comes from the scan or
from a recorded pose of the arm.

\para{Behavior.}
The arm's motion planners, inverse kinematics, collision checks and keep-out checks run against a simulated arm in the
twin. Machine states follow the same rules as on the bench. They include the position of the printer's carriage, the
tool on the carriage, the pick-and-place interlock and the board's progress through the build. The pick-and-place job
runs on the simulated machine built into the machine's own software.

\begin{figure*}[t]
\centering
\includegraphics[width=\textwidth]{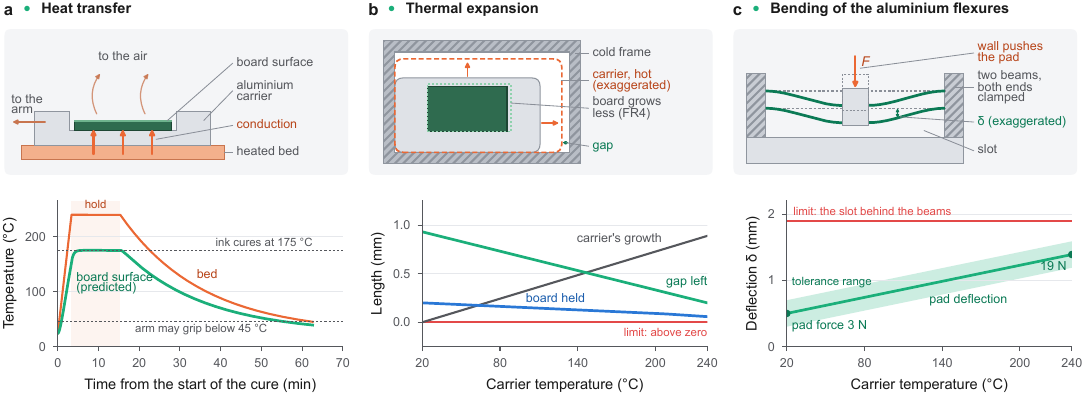}
\caption{\textbf{Physics in the twin.} (a)~Heat transfer. The heated bed warms the aluminium carrier and the
board by conduction; the board loses heat to the air, and the carrier to the arm. The twin runs the cure's own heating
program on a simulated bed and, from the heat path between the bed and the board, predicts the temperature at the
board's surface: it holds at the ink's cure temperature. The twin also predicts when the carrier is cool enough for the
arm to grip. (b)~Thermal expansion. Aluminium grows more than the board and more than the cold frame around the
carrier. Up to 240\,$^\circ$C, a gap stays open around the carrier, and the board stays held. (c)~Bending of the
aluminium flexures. Two beams, clamped at both ends, hold the carrier in place. As the carrier grows, the beams bend
further and push harder; their deflection stays inside the slot behind them. Schematics not to scale; curves computed
by the twin's models.}
\label{fig:physics}
\end{figure*}

\para{Physics.}
The twin predicts what the bench will do and measure (\figref{fig:physics}).
\begin{itemize}
  \item \emph{Heat.} From the bed's temperature and the heat path through the carrier to the board, the twin
  estimates the temperature at the board's surface over time, which sets the curing and reflow profiles. From the materials'
  expansion with temperature, it checks that the carrier's parts still fit together when hot. The carrier holds its
  parts with small aluminium flexure arms; the twin computes how far they bend as the metal grows, and the stress this
  causes, against the material's limit. It also estimates how
  long the carrier needs to cool before the arm may grip it.
  \item \emph{Time.} It estimates the duration of every step, from the arm's motions, the machines' processes and the
  heating and cooling, and so the schedule of the whole run.
  \item \emph{Circuit.} It solves the board's circuit equations, including the part tolerances, the losses of the
  capacitors and the load of the instruments' inputs.
  \item \emph{Measurement.} It holds the calibrated response of the measurement chain, the path from the simulator's
  output through the board to its inputs.
\end{itemize}

\para{Provenance.}
\label{sec:provenance}
Every value in the twin carries its provenance label. The label states how the value was obtained, for example measured,
computed, recorded at a pose of the arm, taken from a vendor, assumed or simulated. It also states
which bench configuration the value is valid for, and which attempt produced it. Every run lists its input files with a
fingerprint (hash) of each. The twin therefore knows not only where things are, but also how it knows. A rehearsal can
then grade each step by the strength of its evidence, and data from different bench configurations can be combined
without confusion.

\section{From a Need to a Rehearsed Build}
\label{sec:plan}

The third stage starts from a need stated in plain language. The agent sets the requirement, turns it into checks,
designs a board that meets them, and rehearses the build of that board in the twin. As an example, we use the 60\,Hz notch filter of the
introduction.

\subsection{From Requirement to Acceptance Code}
\label{sec:spec}

The agent first turns the need into a requirement with numbers, and the requirement into code. The requirement that
the agent set, to remove 60\,Hz mains hum from a $\pm$10\,V signal by at least 40\,dB with a flat passband, becomes
four checks on the board's gain, measured over frequency:
\begin{enumerate}
  \item the \emph{notch}, the frequency of lowest gain, lies within $60\pm1$\,Hz;
  \item the gain at 60.0\,Hz is at most $-40$\,dB;
  \item the passband is flat: far from the notch, at 1\,Hz and at 10\,kHz, the gain is within 0.6\,dB of unity;
  \item the board matches its design: the two $-3$\,dB edges of the notch lie within 10\% of their design values.
\end{enumerate}
The first three checks restate the requirement. Where the requirement states no number, the agent sets one and writes
it down: $\pm$1\,Hz for where the notch may lie, and 0.6\,dB for ``flat'', checked at 1\,Hz and 10\,kHz, far from the
notch on both sides. The fourth check is the agent's own: edges within 10\% of the design show that the board was
built as designed. The agent writes all four as a program, the \emph{acceptance code}. It reads a frequency response,
measured or predicted, and returns a verdict for each check. If the readings do not cover a frequency that a check
needs, the check returns ``cannot evaluate'' instead of a guess. A board passes only when all four checks pass.

\subsection{Design}
\label{sec:design}

The agent then chooses a circuit, sizes it and tests the design statistically. It selects a passive \emph{twin-T}
network, a classic notch circuit of resistors and capacitors. A twin-T needs resistors of $R$, $R$ and $R/2$, and
capacitors of $C$, $C$ and $2C$. The agent uses four resistors of $R=26.4$\,k$\Omega$ (0.1\%) and four capacitors of
$C=100$\,nF (C0G, 1\%); two resistors in parallel make $R/2$, and two capacitors in parallel make $2C$. With these
values, the notch lies at 60.29\,Hz, the gain at 60.0\,Hz is $-52.4$\,dB, and the two $-3$\,dB edges lie at 14.7 and
257.6\,Hz. The agent then uses the twin's circuit model to create 300 virtual boards with random part values within
their tolerances. Every one of them stays below $-43$\,dB at 60.0\,Hz.

Next, programs written by the agent turn the design into everything the bench needs:
\begin{itemize}
  \item the schematic and the board layout, made in KiCad and checked for electrical and design-rule errors;
  \item the files of the conductor layer and of the paste layer, which the printer reads;
  \item the part positions and the pick-and-place job;
  \item the test model, with the list of test frequencies.
\end{itemize}
The layout follows rules for automated builds. All parts are surface-mount and the board has no holes, so no drilling
is needed. Three fiducials guide the pick-and-place camera. A row of six test pads, each with a gold disc,
gives the test head a place to touch the board. The board therefore carries 14 parts: four resistors, four capacitors
and six gold discs.

Before each print, the printer draws a small \emph{ink-start pattern}, which starts the ink flowing evenly. It centers
this pattern on the rectangle that encloses the circuit. The agent adds two small marks at opposite corners of the
board. They move the center of this rectangle, so the pattern lands on bare board, away from the traces. This rule
came from rehearsal: simulated in the twin, the pattern of the first layout crossed the circuit, and the agent
changed the layout by itself.

\subsection{Rehearsal in the Twin}
\label{sec:rehearsal-method}

\begin{table}[t]
\centering
\caption{The step program of one board.}
\label{tab:steps}
\renewcommand{\arraystretch}{1.1}
\begin{tabular}{@{}p{0.15\columnwidth}p{0.68\columnwidth}r@{}}
\toprule
Stage & Steps & No. \\
\midrule
Survey & scan the bench; confirm every machine and fixture & 1 \\
Load & park the printer's gantry; carrier onto the printer's bed & 2 \\
Probe & flow; probe on; height map of the bare board; probe off & 4 \\
Print & flow; ink dispenser on; start the ink and print the traces; dispenser off; leave the flow; park the gantry & 6 \\
Cure & heat, hold, heater off; wait until cool; set the probing limit for the paste; park the gantry & 4 \\
Paste & flow; probe on; two-point alignment and height map; probe off; flow; paste dispenser on; dispense; dispenser
off; end of the flow & 9 \\
Place & park the gantry; interlock check; carrier into the dock; placement job; interlock check; carrier from the
dock straight onto the printer's bed & 6 \\
Reflow & park the gantry; heat, hold, heater off; wait until cool; reset the probing limit; park the gantry;
carrier into the test nest & 6 \\
Test & test head on; sweep and verdict; test head off & 3 \\
Exchange & tested board pushed out; empty adapter onto the next blank board & 1 \\
\midrule
  & \textbf{Total} & \textbf{42} \\
\bottomrule
\end{tabular}
\end{table}

The agent compiles the build and test of one board into a \emph{step program} of 42 steps (\tabref{tab:steps}). Each
step has a kind, which fixes how the step is run and checked:
\begin{itemize}
  \item an arm task: a carrier move, a tool change, the test head, or the board exchange;
  \item a flow of the printer's application, such as a print;
  \item a heating of the bed, or a wait until the bed is cool enough to touch;
  \item a check of the pick-and-place interlock, or a placement job;
  \item the survey, or the test sweep.
\end{itemize}

The rehearsal executes every step in the twin, before any material is used. It covers the experiment from
end to end.
\begin{itemize}
  \item \emph{Motion.} Each arm task is planned and checked by the same script that drives the real arm, run on the
  simulated arm. The rehearsal checks that every pose is reachable, that every path stays clear of the machines and
  their keep-out boxes, and that every grip and set-down meets its fixture.
  \item \emph{Heat.} The thermal model predicts the board's temperature during curing and reflow, checks that the hot
  carrier still fits together, and gives the cooling time that each heating needs before the arm may grip.
  \item \emph{Time.} The durations of all steps give the schedule of the whole run.
  \item \emph{Circuit and test.} The circuit model predicts the board's response at the exact frequencies of the
  planned sweep, and the acceptance code judges this prediction. The rehearsal also checks that the test signals stay
  within their limits.
  \item \emph{Machine files.} The printer's ink-start pattern must land on bare board, and every part must suit the
  nozzle that places it.
\end{itemize}

Each step is then graded by the strength of its evidence:
\begin{itemize}
  \item \emph{proven}: the agent has seen this command run on the bench as written;
  \item \emph{from records}: it rests on facts in the machine's records;
  \item \emph{simulated}: it rests on the twin's simulation alone;
  \item \emph{assumed}: it rests on a value that has not been measured;
  \item \emph{missing}: something it needs does not exist yet;
  \item \emph{at risk}: the rehearsal found a defect in it.
\end{itemize}
Every defect is fixed before execution: in the design, in a program or in the test plan. The real build starts only
when no step is missing or at risk, and no motion rests on an assumed value. The rehearsal also gives the schedule of the run, step by step; most of it is the machines' own heating and cooling.

\section{The Run, Step by Step}
\label{sec:execution}

In the fourth stage, the agent runs the rehearsed step program on the real bench, measures the board, and learns from
the measurement. The run starts with one sentence that a person types on the thinking screen. From then on, the agent
works through the steps by itself.

\begin{figure*}[t]
\centering
\includegraphics[width=\textwidth]{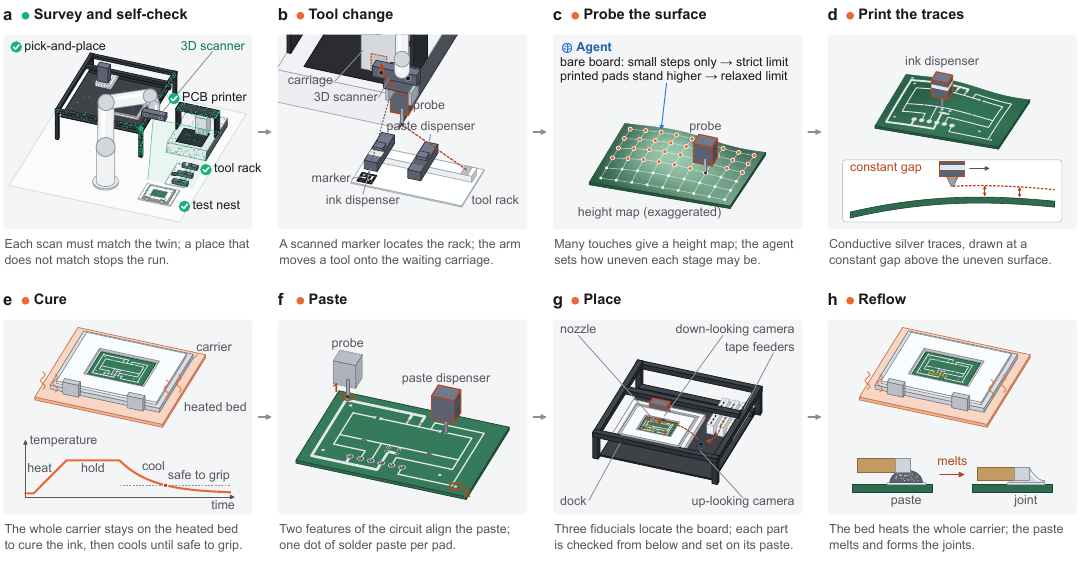}
\caption{\textbf{The build, step by step.} The board is the agent's own layout, shown as it is after each step.
(a)~Before the run, the 3D scanner on the arm's wrist scans every machine and fixture that the run uses, and the agent
matches each scan to its place in the twin; a place that does not match stops the run. (b)~The printer's
carriage waits at its tool-change position. A printed marker, found by the 3D scanner, locates the tool rack, and the
arm moves a tool head from the rack onto the carriage. (c)~The probe touches the bare board at many points. The result is a
height map (unevenness exaggerated), so the nozzle can follow the surface. The nozzle follows the map, so an uneven
board is printed at a constant gap; the agent decides how large a step between neighbouring points each stage may
have, small on a bare board and larger once printed pads stand higher. (d)~The ink dispenser draws the traces in
ink, at a constant gap above the uneven surface (inset: a section along one trace). (e)~The whole
carrier stays on the heated bed while the ink cures, then cools until the arm can grip it safely. (f)~The probe finds
two features of the printed circuit (1, 2), and the paste layer is aligned to them; the paste dispenser sets a dot of
solder paste on each pad. (g)~The down-looking camera finds the board's three fiducials. The nozzle picks each part
from a tape feeder, the up-looking camera checks it, and the part is set on its paste. (h)~The bed heats the whole
carrier again; the paste melts and forms the joints (inset: one joint before and after). Schematic, not to scale.}
\label{fig:process}
\end{figure*}

\subsection{Build}
\label{sec:build}

\para{Survey.}
The arm first scans every machine and fixture that the run uses (\figref{fig:process}a). The agent confirms each one at
its place in the twin. Only then does any part of the bench move under the step program.

\para{Probe.}
The arm carries the carrier, with a blank board in its adapter, onto the printer's bed. It takes the probe from the tool
rack and seats it on the printer's carriage (\figref{fig:process}b). The probe touches the bare board at many points,
and the printer builds a height map of its surface (\figref{fig:process}c). The nozzle will follow this map, so
it keeps the same small gap above a bent or uneven board. The limit for this stage stops only a board whose
neighbouring points differ too much, the sign of a board that is not seated in its adapter, before any ink is spent.

\para{Print and cure.}
The arm swaps the probe for the ink dispenser. The printer draws its ink-start pattern on bare board and prints the
traces in ink. The nozzle follows the height map, so it keeps the same small gap above the slightly
uneven surface (\figref{fig:process}d). The arm removes the dispenser, and the bed heats the whole carrier to cure the
ink (\figref{fig:process}e). The agent waits until the bed is cool enough to grip. Meanwhile, it reasons about the board's
new shape: the printed traces and pads now stand above the bare surface, so the next height map will show larger
differences that are expected. It relaxes the probing limit by that much before the paste.

\para{Paste.}
The arm mounts the probe again. The probe finds two features of the printed circuit, and the printer aligns the paste
layer to them; it also maps the surface again. The arm mounts the paste dispenser, and the printer sets a dot of solder
paste on every pad (\figref{fig:process}f). The paste is a low-temperature tin--bismuth alloy that melts at
138\,$^\circ$C.

\para{Place.}
The arm checks the pick-and-place interlock and sets the carrier into the machine's dock. The machine's down-looking
camera finds the three fiducials and so measures where the board lies and how it is turned. The nozzles take each part
from its tape feeder, the up-looking camera checks it, and the machine sets it on its paste (\figref{fig:process}g).
When the interlock reports the machine at rest again, the arm takes the carrier out.

\para{Reflow.}
The arm brings the carrier back to the printer's bed. The bed heats the whole carrier until the paste melts and forms
the joints, then switches off (\figref{fig:process}h). The thermal model of the twin sets this profile, so that the
board's surface rises well above the melting point of the paste. When the bed is cool enough, the arm carries the
finished board to the test station. The agent also resets the probing limit for the next blank board.

\subsection{Test}
\label{sec:test}

\begin{figure}[t]
\centering
\includegraphics[width=\columnwidth]{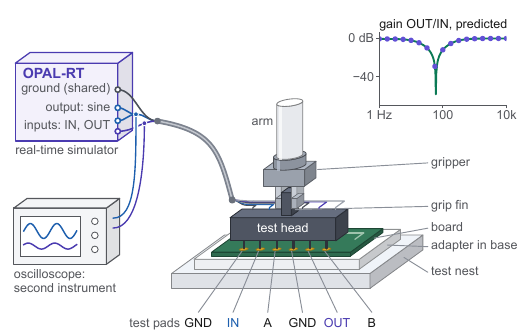}
\caption{\textbf{The hardware-in-the-loop test.} The board, still in its adapter and base, sits in the test nest. The
arm places the test head on the board and holds it there through the sweep; the arm's gripper holds the head by a fin
on its top, and the head's six contact tips touch the board's six test pads, each wired out. A and B are not used by
this board. An analog output of the simulator drives the board's input (IN) with a sine wave. Two analog inputs
record IN and OUT, and the gain OUT/IN is measured at every frequency. An oscilloscope also measures IN and OUT, as a
second instrument. Inset: the twin's prediction of the gain. Schematic, not to scale.}
\label{fig:station}
\end{figure}

The board stays in its carrier for the test. The carrier sits in the test nest, and the arm takes the test head from
its stand, sets it on the nest and holds it there, so that its contact tips touch the board's test pads
(\figref{fig:station}). The head connects the pads to the simulator. One analog output of the simulator
drives the board's input with a sine wave of 1\,V amplitude. The notch board is linear, so its gain does not depend on
the amplitude, and 1\,V keeps the signals well inside the simulator's limits. The same linearity carries the measured
gain to the $\pm$10\,V of the requirement, which the parts withstand: the capacitors are rated for 50\,V, and at 10\,V
no resistor dissipates more than 4\,mW of its 100\,mW rating. Two analog inputs record the board's input
(IN) and output (OUT).

At each frequency, the test model waits until the response has settled. It then compares both recorded signals with a
sine and a cosine at the test frequency over whole periods. This gives the complex gain OUT/IN. The result does not
depend on the level of the source, and broadband noise averages out. The sweep covers 1\,Hz to 10\,kHz with ten points
per decade. Around the notch it adds extra points: 0.5\,Hz apart from 50 to 70\,Hz, and 0.25\,Hz apart from 57 to
64\,Hz. These extra points came from rehearsal: a plain sweep would have missed the notch. The acceptance code then
returns a verdict for each check and for the board.

The oscilloscope measures IN and OUT at the test station at the same time. At each frequency of the sweep, the agent takes
a capture of both channels, keeps it only if it holds the frequency being swept, fits a sine to each channel and
computes its own value of OUT/IN. The two instruments share
the test head and its cable and measure the same points; each digitizes and processes the signals on its own, so their
agreement shows that the measured response is the real response at the board's pads, not an artifact of either
instrument's acquisition.

When the sweep is done, the arm returns the test head to its stand.

\subsection{Board Exchange}
\label{sec:exchange}

While the agent analyzes the data, the arm already prepares the next board. It lifts the adapter, with the tested
board, out of the base and pushes the board out on the push-out pin. It then sets the empty adapter onto the next blank
board, which clips into it (\figref{fig:carrier}b). The bench is then ready for the next build.

\subsection{Learn: from Data to the Next Design}
\label{sec:learn}

A measured response tells more than pass or fail. The agent reads it check by check and compares it with the twin's
prediction at every frequency. Where the board departs from its design, the agent looks for the cause. It weighs each
candidate with numbers from the data, the design and the records: the values of the placed parts, the resistance of
the printed traces and of the solder joints, leakage across the board, and the load of the instruments. Each kind of
cause leaves its own trace in the response. A notch that is off-center points to the products of resistance and
capacitance. A shallow notch points to an imbalance between the circuit's two branches. A loss in the passband points
to a resistance or a load that the design did not include. The agent fits the twin's circuit model, extended by these
effects, to the measured points. The fit recalibrates the twin, so that its next prediction already includes how this
bench builds boards, and it names the design variables that close the gap.

The agent then states its conclusion and its plan for the next board. It chooses new part values from the standard
series, changes the layout, or, when the type of circuit cannot meet the requirement, changes the circuit. It draws the
next board in KiCad, predicts its response in the recalibrated twin and rehearses its build. The loop is then ready for
its next turn.

\subsection{What People See During the Run}
\label{sec:screens}

\begin{figure*}[t]
\centering
\includegraphics[width=\textwidth]{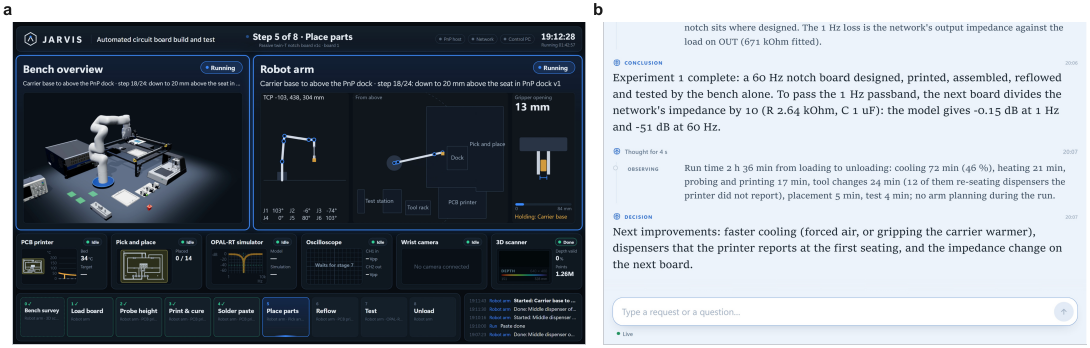}
\caption{\textbf{What people in the lab see during a run}, as recorded in the run of \secref{sec:campaign}.
(a)~The live display, here as the arm carries the carrier to the pick-and-place machine. The bench overview is
the live 3D twin. The device at work is shown large: here the arm, with its pose and joint angles, its place on the
bench seen from above, and what its gripper holds. Every other device stays live in the row below: the printed board
and the bed's temperature, the board waiting for its parts, the simulator and the oscilloscope waiting for the test,
and the 3D scanner after its survey. The run's steps appear along the bottom once the agent has worked them out, and
the current step is highlighted. (b)~The thinking screen at
the end of the run. A person types one request in plain language at the start. The agent shows its reasoning as it
works (what it observes, checks and compares) and states its conclusions and decisions in its own voice; here, the
verdict's explanation, the next design, and the improvements it proposes from the run itself.}
\label{fig:screens}
\end{figure*}

Two screens make the run visible (\figref{fig:screens}). The live display shows the twin as a live mirror of the bench.
It follows the arm's joints, the machines' states and what the gripper holds. Beside it, a tile for each machine shows
its live data: the board as it is printed and populated, the bed's temperature, each part as it is placed, the
measured response drawn over the design curve, and the oscilloscope's waves. The display shows only real data from the
bench. The steps of the run appear along the bottom once the agent has worked them out, and the current one is
highlighted.

The thinking screen shows what the agent thinks, as a researcher would say it: what it observes, how it reads it, what
it checks, what it decides and why. The person who starts the run types the request here, and the agent answers here.
At the end, the agent's conclusion and its plan for the next board appear on this screen. Every message is also kept in
the records, next to the data it refers to.

\section{Results}
\label{sec:experiments}

This section answers four questions behind the claims of this paper.
\begin{itemize}
  \item \textbf{Q1 (Autonomy):} Does \sys{} run the experiment, from a typed request to a tested board and
  the next blank board, without human help?
  \item \textbf{Q2 (The board):} Does the board it built meet the requirement, and do the twin, the simulator and
  the oscilloscope agree on it?
  \item \textbf{Q3 (Learning):} What does the agent conclude from the measured data, and how does it improve the
  design?
  \item \textbf{Q4 (Rehearsal):} What does rehearsal in the twin prevent?
\end{itemize}

\subsection{Q1: The Run Without Human Help}
\label{sec:campaign}

\para{Protocol.}
The agent had set the requirement of \secref{sec:spec} from the need, designed the notch board and run the rehearsal
of its build in the twin.
A blank board lay in the carrier, and the next blank board waited on the table. The agent showed its plan, and the person who had stated the need agreed to it in one sentence
typed on the thinking screen: ``I checked your plan, I agree, pls continue.'' During the run, this was the only input. The agent ran
the step program of \tabref{tab:steps} from the survey to the board exchange. We count the steps completed, the
wall-clock time and the human actions. A human action is any physical or software action by a person between the
request and the verdict. As the laboratory's safety rules require, a person stood by the arm's emergency stop
throughout the run.

\begin{figure*}[t]
\centering
\includegraphics[width=\textwidth]{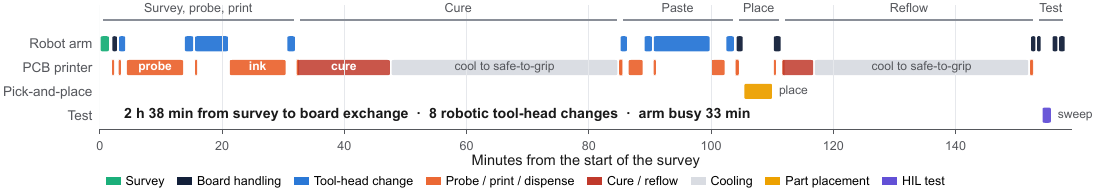}
\caption{\textbf{One board through the bench, as measured in the run.} Each bar is an activity of one machine. The arm carries the board, changes the printer's tool heads
eight times, holds the test head and exchanges the board. Most of the time is the machines' own process time: curing,
reflow and the cooling after each.}
\label{fig:timeline}
\end{figure*}

\para{Results.}
\sys{} completed all 42 steps in 2\,h 39\,min, from the request to the next blank board, with zero human actions
(\figref{fig:timeline}). The survey confirmed the test station, the carrier and the pick-and-place dock at their
places in the twin. The probe mapped the board before both deposits; the pick-and-place machine found the three
fiducials and placed all 14 parts; and the heating held the bed at 240\,\textdegree{}C for 12\,min to cure the ink and
for 2\,min to reflow the paste, and released the carrier only when it was cool enough to grip. This was the first time
the experiment ran from end to end. As in any first run on real machines, it met faults of the
kinds common in computer and engineering practice: a tool head that the printer did not recognize once seated, and a
control of the printer's application that stopped responding. Each time, the agent read the machine's own state, chose
a remedy and went on. It had the unrecognized head taken off and seated again, after sending the printer's carriage
home where needed, until the printer reported it. When the control that positions the circuit on the board stopped
responding, the agent read the circuit's position, found it within 0.2\,mm of its target, accepted it and went on. No
person stepped in. Every arm task ran on a plan checked in the twin before the run, so the arm never stopped to plan. The arm's
work, including the eight tool changes, the test head and the board exchange, took 33\,min of the run. The rest was
the machines' own process time; heating and cooling alone took 93\,min. The run ended with the tested board out of its
adapter and the next blank board loaded, ready for the next build. The run is itself one attempt in the ledger, whose log records every step, so every
step can be traced from the request to the verdict and the next design.

\subsection{Q2: The Board Against the Requirement}
\label{sec:board}

\begin{figure}[t]
\centering
\IfFileExists{figures/board-photo.jpg}{\includegraphics[width=\columnwidth]{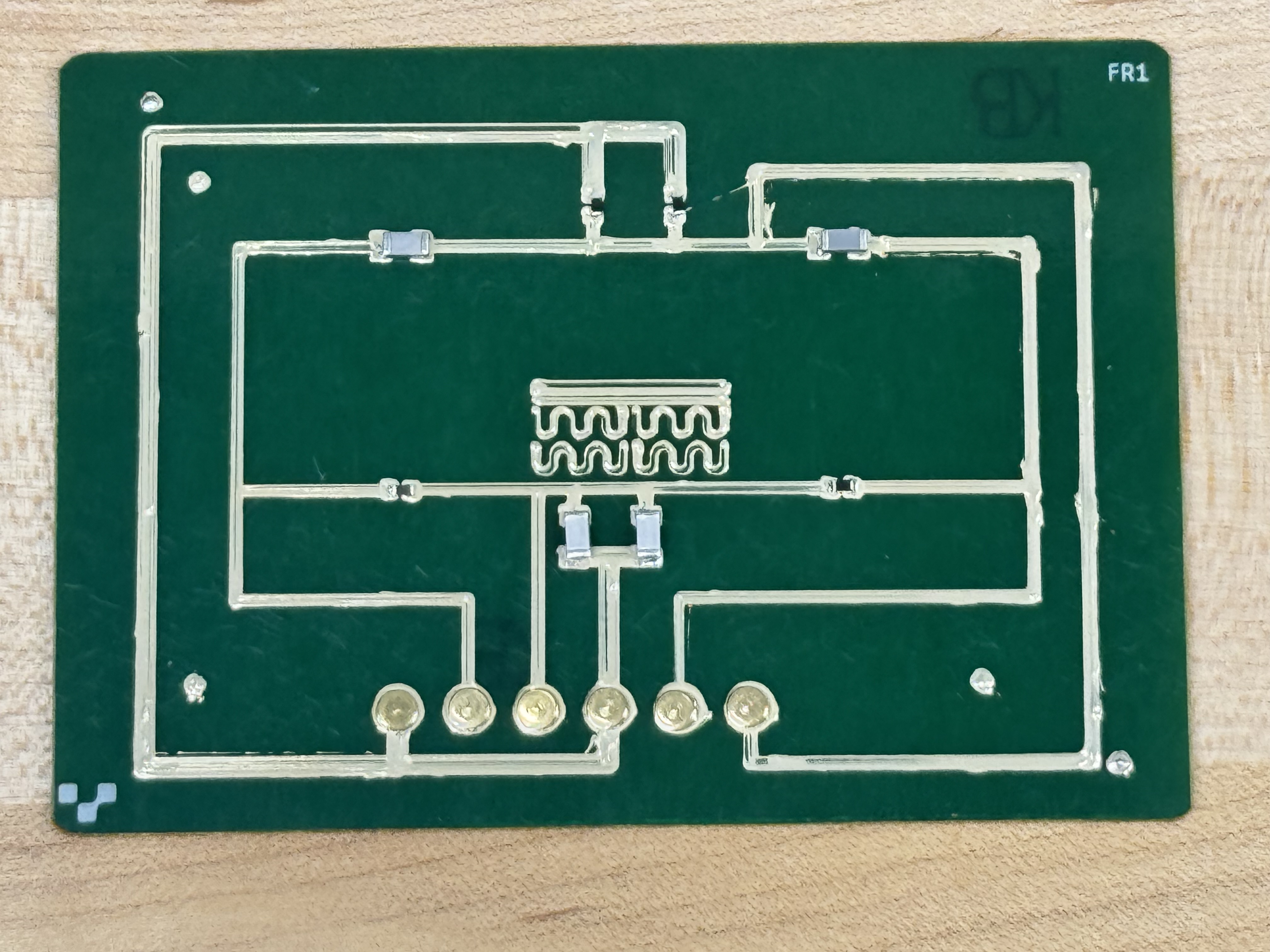}}{%
\plannedpanel{Photo of the board built and tested in the run (figures/board-photo.jpg).}}
\caption{\textbf{The board built and tested by \sys{}.} The traces printed in ink, the 14 parts placed
and reflowed, and the gold discs on the test pads, as the board came out of the run.}
\label{fig:boardphoto}
\end{figure}

\begin{figure}[t]
\centering
\includegraphics[width=\columnwidth]{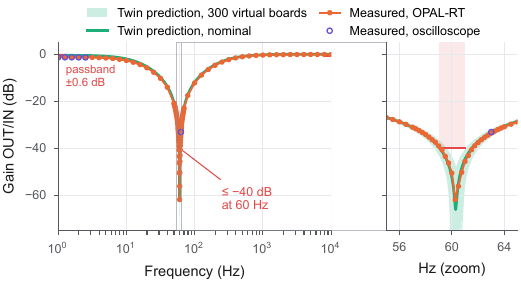}
\caption{\textbf{The requirement as code, the twin's prediction and the measured board.} Red: the acceptance limits
derived from the requirement. Green: the twin's predicted response of the notch board, for nominal parts and for 300
virtual boards with random part values within their tolerances. Orange: the board measured by the simulator; open circles: the oscilloscope's readings.}
\label{fig:response}
\end{figure}

\begin{table}[t]
\centering
\caption{The board's verdict, check by check.}
\label{tab:verdict}
\setlength{\tabcolsep}{3.5pt}
\begin{tabular}{@{}lcccc@{}}
\toprule
Check & Limit & Twin & Measured & Verdict \\
\midrule
Notch frequency & $60\pm1$\,Hz & 60.29\,Hz & 60.29\,Hz & pass \\
Gain at 60\,Hz & $\le-40$\,dB & $-52.4$\,dB & $-50.6$\,dB & pass \\
Passband, 1\,Hz & $\pm0.6$\,dB & $-0.47$\,dB & $-1.29$\,dB & fail \\
Passband, 10\,kHz & $\pm0.6$\,dB & $-0.02$\,dB & $-0.16$\,dB & pass \\
Lower $-3$\,dB edge & 14.7\,Hz $\pm$10\% & 14.7\,Hz & 15.3\,Hz & pass \\
Upper $-3$\,dB edge & 257.6\,Hz $\pm$10\% & 257.6\,Hz & 263.8\,Hz & pass \\
\bottomrule
\end{tabular}
\end{table}

\para{Protocol.}
The simulator swept the board as described in \secref{sec:test}, and the oscilloscope measured it at the same time.
The acceptance code judged the simulator's sweep. We compare it with the twin's prediction made before the build, and
with the oscilloscope's readings.

\para{Results.}
\figref{fig:boardphoto} shows the board as it came out of the run. \figref{fig:response} and
\tabref{tab:verdict} show it against its requirement. The notch, located by a
parabola through the three lowest readings, lay at 60.29\,Hz, the frequency that the twin had predicted (the deepest
reading was at 60.25\,Hz), and the board removed 50.6\,dB at 60\,Hz, 10.6\,dB more than the
requirement asks. The edges of the notch lay within 4.0\% of their design values, so the board was built as designed.
The passband was flat at high frequency, $-0.16$\,dB at 10\,kHz. At 1\,Hz the gain was $-1.29$\,dB, 0.69\,dB beyond
its limit, where the twin had predicted $-0.47$\,dB with the simulator's 1\,M$\Omega$ input as the only load
($-0.51$\,dB with the oscilloscope's 10\,M$\Omega$ as well); \secref{sec:learn-results}
traces this difference to its cause. At every frequency that the oscilloscope read, it agreed with the simulator: to within
0.05\,dB in the passband and to within 0.43\,dB on the steep flank of the notch at 63\,Hz. The measured response is
therefore real, not an artifact of either instrument.

The oscilloscope read only six of the sweep's 96 frequencies, a flaw of the agent's test program. Its captures, with
their transfer to the control computer, took longer than the simulator's points above a few hertz, and the program did
not wait for them, so most captures did not hold the frequency being swept and were discarded. The agent traced this
cause in the run's own log, and its next test program will hold each frequency until the oscilloscope has read it. Subsequently, we conducted a re-measuremen, and at every frequency that the oscilloscope read, it agreed with the
simulator.

\subsection{Q3: From the Measured Data to the Next Design}
\label{sec:learn-results}

\begin{figure}[t]
\centering
\includegraphics[width=\columnwidth]{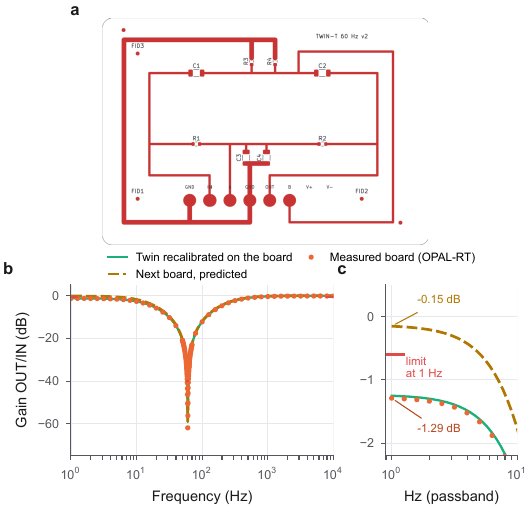}
\caption{\textbf{The next design.} (a)~The board that the agent proposed from the measured data, as drawn in KiCad (copper and
silkscreen): the same layout, every resistance divided by ten and every capacitance multiplied by ten. (b)~The measured board, the twin
recalibrated on it, and the next board predicted by the recalibrated twin. (c)~The passband near 1\,Hz enlarged: the
next board meets the limit that the built board missed.}
\label{fig:nextboard}
\end{figure}

\para{Protocol.}
After the verdict, the agent analyzed the measured response by itself, as described in \secref{sec:learn}. It
displayed its reasoning on the thinking screen and wrote its conclusion and its plan for the next board into the
records.

\para{Results.}
The agent traced the low-frequency sag to its cause. It weighed the candidates against the data. Errors in the part values would move the notch or its edges, but the notch lay at
the twin's predicted frequency and both edges within 4.0\% of their design values, as the 0.1\% resistors and 1\%
capacitors lead one to expect. A poor solder joint would make
the notch shallower, but the notch was 50.6\,dB deep. What remains is the network's output impedance against the load
on its output: it lowers the gain at low frequency only, where that impedance is highest. The fit of the extended
circuit model to the 96 measured points recalibrated the twin, which then reproduced the measured response to within
0.14\,dB (rms). The data fix one quantity: the network's output impedance relative to the load on its output, which the
fit puts at 2.6 times the design's ratio. With the parts at their nominal values, this is a load of about
340\,k$\Omega$ on the output, against the 909\,k$\Omega$ assumed for the inputs of the two instruments. Two sources can
account for the difference: the simulator's input, whose documentation gives 500\,k$\Omega$ to ground as well as
1\,M$\Omega$ between its terminals, and leakage across the board. Whichever carries it, the remedy is the same. From this, the agent
proposed the next board: the same circuit with every resistance divided by ten and every capacitance multiplied by ten
(2.64\,k$\Omega$ and 1\,$\mu$F). This keeps every RC product, and so the notch, and makes the network ten times stiffer
against the load. It also loads the source ten times more: at 10\,kHz the network's input impedance falls to about
660\,$\Omega$, so a 10\,V signal draws 15\,mA, the rating of the simulator's analog outputs, and the 1\,V test signal
1.5\,mA. It drew the board in KiCad (\figref{fig:nextboard}), and the recalibrated twin predicts $-0.15$\,dB
at 1\,Hz, inside the limit, with $-51.0$\,dB at 60\,Hz.

\para{Proposals beyond the design.} The agent did not stop at the verdict. From the record of its own run, it also
proposed how to run the next experiment better. Cooling the bed after each heating took nearly half of the run, so it proposed
forced-air cooling of the bed, or gripping the carrier at a higher temperature where its materials allow. Tools that
their machine did not recognize at once had cost minutes, so it proposed a firmer seating of the tools. A conventional
automated platform executes a fixed recipe and reports its result. \sys{} explains the result and proposes how to
improve both the design and the experiment that tests it. A single run thus gave the system a
verdict, an explanation and a concrete next step, without a person interpreting the data.

\subsection{Q4: What Rehearsal in the Twin Prevents}
\label{sec:rehearsal}

\begin{table}[t]
\centering
\caption{Defects that rehearsal in the twin exposed before any board was built, and the agent's fix for each.}
\label{tab:rehearsal}
\renewcommand{\arraystretch}{1.12}
\begin{tabular}{@{}p{0.47\columnwidth}p{0.45\columnwidth}@{}}
\toprule
Defect exposed in the twin & Fix by the agent \\
\midrule
\textbf{Test plan.} The first sweep had 25 points, evenly spaced in log-frequency from 10\,Hz to 1\,kHz. It had no
point between 56.2 and 68.1\,Hz, so its deepest reading would be $-29$\,dB. Even a correct board would fail check 2.
& A sweep from 1\,Hz to 10\,kHz, with extra points around the notch down to 0.25\,Hz apart. The predicted verdict
becomes a pass. \\
\textbf{Ink-start pattern.} The printer centers this pattern on the rectangle that encloses the circuit. 120 of its 548
strokes would cross the traces. & Two small marks move the center. The pattern then lands on bare board
(\figref{fig:dfa}). \\
\textbf{Paste alignment.} The printer places the paste layer from two probed points and the layers' enclosing
rectangles. The conductor layer and the paste layer must therefore have the same rectangle. & The same two marks on
both layers. \\
\textbf{Placement nozzle.} The 1206 capacitors are 1.6\,mm wide, below the minimum part size of the default nozzle. &
The capacitors are assigned to the finer nozzle. \\
\textbf{Keep-out box.} The keep-out box around the printer must open for each of the eight tool changes and stay
closed otherwise. & The box follows the printer's reported carriage position. \\
\bottomrule
\end{tabular}
\end{table}

\begin{figure}[t]
\centering
\includegraphics[width=\columnwidth]{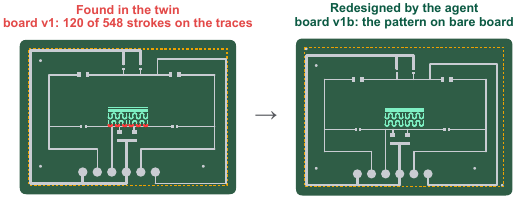}
\caption{\textbf{A defect found by simulation in the twin, and the agent's own redesign.} The printer draws
its ink-start pattern at the center of the rectangle that encloses the circuit. Simulated in the twin, the
pattern on the first layout, v1, would cross the traces (red strokes). No person pointed this out: the agent found it
in the rehearsal and changed the design itself, adding two marks that move the center. On the revised layout, v1b,
whose marks the built board keeps, the pattern lands on bare board.}
\label{fig:dfa}
\end{figure}

\para{Protocol.}
The agent rehearsed the complete build of the first layout (v1) in the twin, as described in
\secref{sec:rehearsal-method}. The rules and settings of \secref{sec:plan} and \secref{sec:execution} already include
the fixes that this rehearsal produced.

\para{Results.}
The rehearsal exposed the five defects of \tabref{tab:rehearsal} before any material was used. Each was found by
simulation in the twin, and each was fixed by the agent itself: it changed the design, a program or the test
plan, and rehearsed again. Each of them would have cost at least one board. The first would have done worse: it would have rejected a correct board and sent the loop to
redesign a circuit that already met its requirement. A loop that learns from measurements must be protected from a
wrong test plan, and rehearsal makes the test plan itself something that is checked. After the fixes, every arm task of
the run, from carrier moves and tool changes to the test head and the board exchange, passed every check of its script
in the twin.

\section{Discussion: Toward Self-Driving Hardware Labs}
\label{sec:discussion}

\para{The twin as a world model.}
A typical digital twin predicts the state of a physical system. The twin of \sys{} also runs the agent's own programs,
so it answers what will happen if the agent acts, and on what evidence that answer rests. Three properties make this possible.
The twin executes the production code, so a rehearsal tests the very program that will run. The agent builds the twin
by scanning the bench, so the twin stays current without a modeling team. Every value carries its provenance, so a
prediction states the evidence it rests on. Together, they make the twin a world model in the full sense: a place to
think before acting, about motion, heat, time and the circuit at once.

\para{An agent that understands its result.}
The run does not end at a verdict. The agent reads the measured response, explains it with the physics of the board
and of the build, and turns the explanation into the next design. This is the step that a self-driving laboratory for
hardware needs most: a new circuit is a new object, so the next build cannot be found by tuning a recipe. It must be
reasoned out from the data, as an engineer would do it.

\para{What the system enables next.}
The same loop opens several directions.
\emph{More turns of the loop:} the next board that the agent drew will be built by the same step program, and each turn
recalibrates the twin with how this bench builds boards.
\emph{New requirements:} a 50\,Hz grid needs only new part values; a narrower notch calls for an active circuit, for
which the agent changes the circuit model of the twin and the test plan, and the rest of the loop stays as it is.
\emph{Design-space exploration:} the twin can rehearse and predict many candidate designs, so the bench builds only the
most informative board, in the spirit of Bayesian experimental design~\cite{shields2021bayesian}.
\emph{Learning:} every rehearsed and executed task leaves a labeled trajectory with its provenance. Such trajectories
are training data for learned manipulation policies and for real-to-sim transfer~\cite{torne2024rialto,
kerbl2023gaussian}.
\emph{Throughput:} the arm is free during most of each build, while the board cures, reflows and cools, so one arm can
serve several printers and test stations, with the twin planning their shared use.

\para{Generality.}
Few parts of \sys{} are specific to notch filters: the circuit model in the twin's physics layer, the acceptance code
and the test plan. A new circuit family needs its own versions of these; the rest of the system stays the same. A new
machine joins the bench as described in \secref{sec:agent}: it gets its records, its scans and its skills, and the
rehearsal then covers it like any other machine. The next step within power electronics is power stages and embedded
controllers, tested hardware-in-the-loop against simulated grids and machines. This is the use for which real-time
simulators are built~\cite{faruque2015realtime}, and the bench already has one.

\para{Design choices.}
Three choices make \sys{} reproducible and trustworthy. Operating machines through the interfaces that come with them
keeps every machine standard, so another laboratory can rebuild the bench from equipment of the same kinds. Checking
every step with the system's own sensors, and measuring with two instruments, makes the result stand
without a person's inspection. Logging every attempt, and showing the agent's reasoning as it works, makes the agent's
physical work as reviewable as its code.

\section{Conclusion}
\label{sec:conclusion}

\sys{} closes the hardware design loop with an embodied agent. A need stated in plain language goes in. A circuit
board comes out: designed, built on unmodified machines and tested hardware-in-the-loop, without a human hand
in between, together with the agent's explanation of what it measured and its design for the next board. The agent
gets there in four stages. It knows its machines through records that state the source of every fact. It scans the
bench and builds its own executable digital twin, in which it rehearses the experiment before any material is used. It
runs the experiment on the bench, checking every step with its 3D scanner and the machines' own cameras and sensors,
and measuring the board with two instruments. And it learns from the data: it explains the result and turns
it into the next design. A ledger accounts for every physical action, and two screens let people watch the bench and
the agent's reasoning as they work. Together, these stages turn a lab bench into a self-driving laboratory for
electronic hardware.

\section*{A Personal Perspective}
\label{sec:perspective}

Speaking as researchers in electrical and computer engineering, we find it remarkable that an agent, on its very first
autonomous attempt, carried this experiment through so smoothly and so completely: from a need stated in a sentence
to a working board that passes nearly every check, resolving the unexpected on its own along the way. A first run
on real machines rarely goes this well, even in the hands of an experienced engineer.

We also see clearly how far there is still to go. The board bears the mark of a capable newcomer rather than a seasoned
designer. Its traces turn at right angles, where established practice favors 45-degree bends; at 60\,Hz this costs
nothing electrically, but it is not how a professional would lay out a board. Some traces detour along the board's edge
instead of taking the shortest path, and the impedance level of the network proved too high for the load it drives,
which is what cost the 1\,Hz passband. Each of these is the kind of judgment that experience brings, and that an
autonomous designer has yet to acquire. In our view, the agent's analysis of the measurement and its new design are
not perfect either, and we hope to improve them.

We therefore see this experiment as a beginning rather than an end. Our aim is to grow this fully automated loop of
design, fabrication and test into a platform of professional quality, and to contribute what it teaches us to the
research community, so that others can build on it and help bring self-driving laboratories to electronic hardware.


\end{document}